\pdfoutput=1

\documentclass[11pt]{article}

\usepackage[preprint]{acl}

\usepackage{times}
\usepackage{latexsym}

\usepackage{pifont}
\definecolor{darkgreen}{HTML}{2D8A46}
\newcommand{\checkmark}{\textcolor{darkgreen}{\ding{51}}}
\newcommand{\xmark}{\textcolor{red}{\ding{55}}}

\usepackage{makecell}

\usepackage{tabularx}
\newcolumntype{Y}{>{\centering\arraybackslash}X}
\renewcommand{\arraystretch}{1.6}

\usepackage{array}
\usepackage{booktabs}
\usepackage{amsmath}

\usepackage{adjustbox} 

\newcommand{\figref}[1]{Fig.~\ref{#1}}

\newcommand{\eqnref}[1]{Eq.~\eqref{#1}}

\newcommand{\tabref}[1]{Tab.~\ref{#1}}

\newcommand{\boldparagraph}[1]{\vspace{0.03cm}\noindent{\bf #1:}}

\usepackage{listings}
\usepackage{xcolor}

\definecolor{codegreen}{rgb}{0,0.6,0}
\definecolor{codegray}{rgb}{0.5,0.5,0.5}
\definecolor{codepurple}{rgb}{0.58,0,0.82}
\definecolor{backcolour}{rgb}{0.95,0.95,0.92}
\lstdefinestyle{promptstyle}{
    backgroundcolor=\color{backcolour},   
    commentstyle=\color{codegreen},
    keywordstyle=\color{magenta},
    numberstyle=\tiny\color{codegray},
    stringstyle=\color{codepurple},
    breakatwhitespace=false,         
    breaklines=false,                 
    captionpos=b,                    
    keepspaces=true,                 
    numbers=left,                    
    numbersep=5pt,                  
    showspaces=false,                
    showstringspaces=false,
    showtabs=false,                  
    tabsize=2,
    basicstyle=\tiny\ttfamily
}

\usepackage[T1]{fontenc}

\usepackage[utf8]{inputenc}

\usepackage{microtype}

\usepackage{inconsolata}

\usepackage{graphicx}

\title{\raisebox{-0.05cm}{\includegraphics[scale=0.13]{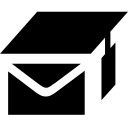}} Scholar Inbox: Personalized Paper Recommendations for Scientists}

\author{Markus Flicke\quad Glenn Angrabeit\quad Madhav Iyengar\quad Vitalii Protsenko
\vspace{-0.2cm}\\ {\bf Illia Shakun\quad Jovan Cicvaric\quad Bora Kargi\quad Haoyu He\quad Lukas Schuler} 
\vspace{-0.2cm}\\{\bf Lewin Scholz\quad Kavyanjali Agnihotri\quad Yong Cao\quad Andreas Geiger}\\
University of Tübingen, Tübingen AI Center\\
\url{www.scholar-inbox.com}
}

\begin{document}
\maketitle

\begin{abstract}
Scholar Inbox is a new open-access platform designed to address the challenges researchers face in staying 
current with the rapidly expanding volume of scientific literature.  We provide personalized recommendations, 
continuous updates from open-access archives (arXiv, bioRxiv, etc.), visual paper summaries, semantic search, 
and a range of tools to streamline research workflows and promote open research access.  
The platform's personalized recommendation system is trained on user ratings, 
ensuring that recommendations are tailored to individual researchers' interests.  
To further enhance the user experience, Scholar Inbox also offers a map of science that provides an 
overview of research across domains, enabling users to easily explore specific topics.   
We use this map to address the cold start problem common in recommender systems,
 as well as an active learning strategy that iteratively prompts users to rate a selection of papers, 
 allowing the system to learn user preferences quickly.  
We evaluate the quality of our recommendation system on a novel dataset of 800k user ratings, 
which we make publicly available, as well as via an extensive user study.
\end{abstract}

\section{Introduction}
\begin{figure}[!ht]
    \centering
    \includegraphics[width=0.49\textwidth]{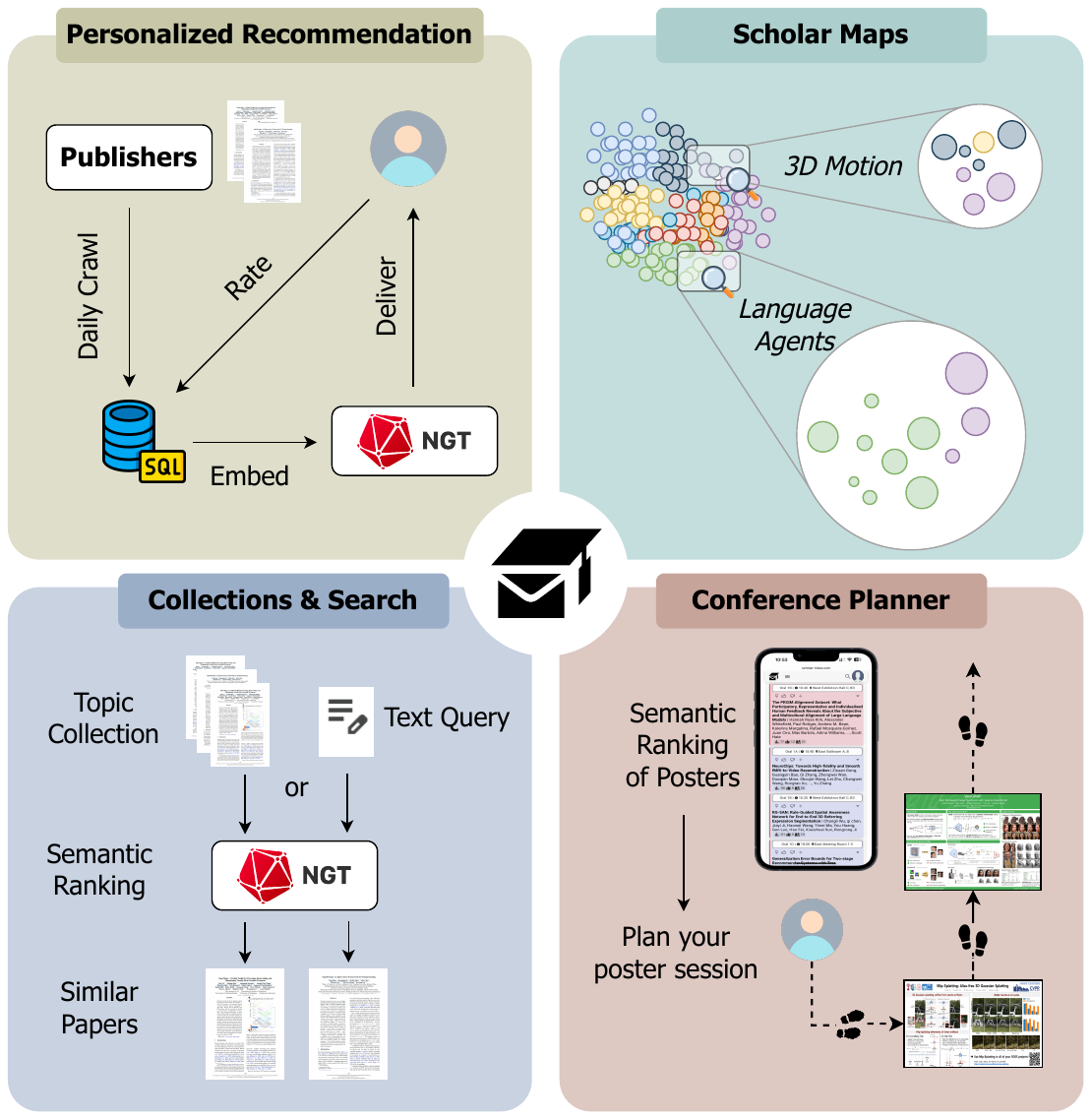}
    \caption{
         Key features of Scholar Inbox, including \textit{Personalized Recommendations} tailored to individual interests, 
         \textit{Scholar Maps} for cross-domain  paper exploration, \textit{Collections} 
         for literature review and exploration of new research areas, and \textit{Conference Planner} for
         efficient time prioritization at conference poster sessions.\\
    }
    \label{fig:key_features}
\end{figure}
The exponential growth of scientific publications has posed significant challenges for both junior and senior researchers to stay up-to-date 
with the latest relevant works \cite{fortunato_science_2018, zheng-etal-2024-openresearcher}. 
This motivated the development of academic recommenders, which offer personalized paper recommendation services, 
aiming to promote the discovery of relevant works and enhance the efficiency of the research cycle.

However, despite these efforts, current platforms often fail to fully meet user requirements. 
For example, many researchers rely on platforms like X\footnote{\url{www.x.com}}, 
ResearchGate\footnote{\url{www.researchgate.net}} or
LinkedIn\footnote{\url{www.linkedin.com}} for paper recommendations, 
which implicitly introduce biases towards popular authors and institutions via the Matthew effect \cite{perc2014matthew, farber2023biases}. 
Furthermore, where personalized recommendations are offered, they are typically based on broadly defined topics \cite{paperdigest}, 
leading to an inaccurate understanding of user interests and thus suboptimal paper recommendations \cite{li2021personalized}.

In this paper, we present Scholar Inbox, a publicly available open-access platform with more accurate personalized recommendations and 
a wide range of functionalities for researchers, aiming to enhance research efficiency and promote open-access publications. 
As shown in \figref{fig:key_features}, the advantages of Scholar Inbox primarily include four aspects: 
\boldparagraph{(1) Personalized Recommendations} We train a recommendation model for each 
researcher based on their positive and 
negative ratings during registration and 
 while visiting our website. 
Unlike social media recommendations, our recommendations are only based on the paper content and therefore unbiased by social factors. 
\boldparagraph{(2) Scholar Maps} To facilitate exploration of papers across domains, 
we project all papers into a two-dimensional space based on their semantic representations, 
allowing users to easily search and discover research.
\boldparagraph{(3) Collections and Search} We enable users to explore papers that are semantically similar 
to their collections and search similar papers based on free-form text descriptions.
\boldparagraph{(4) Conference Planner} For large conferences, we offer a planner that helps users prioritize their time at poster sessions.  

Besides offering a range of functionalities, we introduce a content-based recommendation model for research papers, 
provide a demonstration video\footnote{\url{https://youtu.be/4fgM-iJgXJs}},
 and release our dataset\footnote{\href{https://www.github.com/avg-dev/scholar_inbox_datasets}{github.com/avg-dev/scholar\_inbox\_datasets}}
 of anonymized user ratings to support and facilitate future research on scientific recommender systems.
In the following sections, we summarize existing academic platforms ($\S$\ref{sec:related_work}), present the system architecture of Scholar Inbox
 ($\S$\ref{sec:scholar_inbox}), and provide comprehensive evaluations, 
 demonstrating its ability to deliver better recommendations and enhance user satisfaction  ($\S$\ref{sec:evaluation}).


\section{Related Work}
\label{sec:related_work}
\boldparagraph{Scientific Paper Recommendation Platforms}
To meet growing research demands, support systems such as search engines, exploratory tools, and recommenders have emerged.
Search engines like Google Scholar and Semantic Scholar rely on user-provided keywords.
Research interests are however often multi-faceted and many new researchers are unaware of which terms accurately describe 
their desired search results.
Exploratory tools such as Connected Papers\footnote{\url{www.connectedpapers.com}} and 
Research Rabbit\footnote{\url{www.researchrabbit.ai}} fill this gap by visualizing 
citation graphs as 2D maps to show related papers to the user. 
Additionally, semantic paper maps of research have been created using t-SNE \cite{pmlr-v196-gonzalez-marquez22a}.
\\\boldparagraph{Recommendation Algorithms}
Beyond exploration, researchers must read the latest research to stay relevant in their field and to avoid duplicate research. 
A plenitude of research recommenders have been proposed, 
but no system has so far achieved widespread adoption. 
Content-based filtering (CB) recommendation systems \cite{arxivsanity, wang_content-based_2018, patra_content-based_2020, kart_emati_2022} 
generate recommendations purely using item information, but have been refined to include user interactions 
\cite{mohamed_tag-aware_2022, guan_document_2010} and bibliographic information \citep{ma_chronological_2021, 
wang_content-based_2018}. 
Many implementations prefer sparse Term Frequency Inverse Document Frequency (TF-IDF) \cite{jones_statistical_1972} 
embeddings over dense learning-based embeddings, 
due to their simplicity and lower runtime
\cite{zhang_scholarly_2023,MohamedHassan20196}. 
Our ablation study corroborates that TF-IDF performs well for the research recommendation task, 
however we find that state-of-the-art distributed representations such as GTE \cite{li2023towards} 
outperform sparse embeddings in terms of vote prediction accuracy. 
\\
A known limitation of CB recommendation systems is the filter bubble effect \cite{portenoy_bursting_2022} and diversity, 
novelty and serendipity have been identified as current limitations \cite{kreutz_scientific_2022, ali_overview_2021, 
bai_scientific_2019, nguyen_exploring_2014}. 
In contrast, collaborative filtering (CF) derives recommendations from multiple users' interests and current approaches differ by 
whether they utilize author information \cite{utama_scientific_2023,neethukrishnan_ontology_2017}, use interactions 
\cite{murali_collaborative_2019, xia_folksonomy_2014} or bibliographic information 
\cite{sakib_collaborative_2020,haruna_collaborative_2017,liu_context-based_2015}. 
\\
Recent work focuses on hybrid systems, incorporating CB and CF into two-tower architectures 
\cite{church_academic_2024, yi_sampling-bias-corrected_2019} or graph based approaches 
\cite{wang_multi-feature-enhanced_2024,ostendorff_neighborhood_2022,cohan_specter_2020}. 
CB, CF and hybrid approaches all suffer from the cold start problem for recommendation systems, 
as the recommender is uninformed about user preferences when they begin to use the system \cite{bai_scientific_2019}. 
There have been many attempts to alleviate this problem \cite{nura_author-centric_2024}, 
for instance by uploading bibtex files from a reference manager \cite{kart_emati_2022}. 
Scholar Inbox mitigates the cold start problem through a user-friendly onboarding process and an active learning strategy.
\\\boldparagraph{Research Recommendation Datasets}
There are only a few research recommendation datasets available, such as Semantic Scholar Co-View \cite{cohan_specter_2020}, 
SPRD \cite{sugiyama_scholarly_2010} and the largest dataset, 
CiteULike \cite{wang_collaborative_2011}, contains 205k interactions. 
CiteULike's user-paper interaction are made when a user assigns a paper to their library, 
which implicitly shows that they liked that paper, 
but the exact reason why they added this paper is unclear. 
There is a lack of standard datasets in the field \cite{sharma_anatomization_2023}, 
which is the reason we are releasing a dataset of 800k explicit positive/negative rating interactions
from over 14.3k users.
Furthermore, studies analyzing users' feedback to improve scholarly recommendation systems are rare and  
have very low number of responses \cite{zhang_scholarly_2023}. 
We describe the outcomes of our user study with over 1.1k participants in the evaluation section.

\section{Scholar Inbox}
\label{sec:scholar_inbox}
Our proposed scientific paper recommender system contains several key features, 
which we order by popularity according to our user survey: 
\\\boldparagraph{Daily Digest}
Daily paper updates (\figref{fig:digest}), ranked according to user interests provide 
a systematic way to keep up to date with research in the user's area of focus.
The daily frequency of updates is designed to allow the user to build strong habits around staying informed in research.
\\\boldparagraph{Semantic Search}
Users can search for papers by inserting free-form text. 
Example use-cases are to search for missed citations of related work sections, 
or to find papers that are similar to a paper the researcher is currently working on.
\\\boldparagraph{Conference Planner}
Academic conferences are important for exchanging ideas, staying informed, and networking. 
To support this, we provide a poster session planner for leading machine learning conferences, 
which includes a personalized ranked list of posters and the ability to bookmark papers for later reading. 
We plan to extend this service to all scientific disciplines in the near future. 
\\\boldparagraph{Collections}
Any paper can be added to a user's collection for later reading. 
We show similar papers to each collection, such that the user can exploratively expand their collection. 
\\\boldparagraph{Figure Previews}
Along with the title, abstract and authors, 
we show the first five tables and figures of each paper, which we extract from the paper pdf using papermage \cite{lo_papermage_2023}.

\subsection{Recommendation Model}
To sort papers by relevance, Scholar Inbox uses a content-based recommender, 
which trains a logistic regression model on the user's paper ratings.

\subsubsection{Training}
\label{subsubsec:training}
Unlike traditional recommender systems that rely solely on implicit feedback from item interactions, 
Scholar Inbox enables users to tune their classifier through explicit up and downvotes.
In addition to user ratings, we sample 5k random negative papers that the user has not interacted with,
to better regularize the decision boundary. 
In contrast, our users have an average of 78 positive ratings, leading to a highly imbalanced dataset.
To address this class imbalance, we use the weighted binary cross-entropy loss and assign distinct weights to positive ratings ($w_P$), negative ratings ($w_N$), and randomly sampled negatives ($w_R$):
$$
\mathcal{L}= \frac{1}{n_T}\sum_{i=1}^{n_T} -w_i[y_i\log{\hat y_i}+(1-y_i)\log{(1-\hat y_i)}]
$$
where $n_T$ equals the total training set size. With $n_P$, $n_N$, and $n_R$ representing the number of papers in each group, that is $n_T = n_P + n_N + n_R$, the weights of the two classes are balanced by:
\begin{equation}
n_P ~ w_P = S ~ \left(n_N ~ w_N + n_R ~ w_R \right)
\label{eq:balance}
\end{equation}
While the hyperparameter $S$ controls the overall magnitude of negative weights ($w_N$ and $w_R$), we introduce another hyperparameter $V$ to adjust the relative importance between explicit negative ratings and randomly sampled negatives.
For any chosen value of $V \in [0, 1]$, \eqnref{eq:balance} is then satisfied using the following 
intermediate weights: $\tilde{w}_P = \frac{1}{n_P}$,\\ $\tilde{w}_N = \frac{S ~ V}{V ~ n_N + (1 - V) ~ n_R}$, $\tilde{w}_R = \frac{S~(1 - V)}{V ~ n_N + (1-V) ~ n_R}$.
 
\vspace{6.5pt}
\noindent
This formulation ensures that as users provide more explicit negative votes, the influence of randomly selected negatives on the overall weighting diminishes. However, it introduces a bias in the mean cross-entropy loss. Assuming each sample has an unweighted cross-entropy loss of 1, we derive:
$$
\mathcal{L} = \frac{1}{n_T}
\left(n_P ~ \tilde{w}_P + n_N ~ \tilde{w}_N + n_R ~ \tilde{w}_R\right) = \frac{S+1}{n_T}
$$
This dependency on the total training set size $n_T$ becomes problematic when applying weight decay and tuning the inverse regularization parameter $C$ across users with different numbers of ratings. 
To correct for the bias, we multiply all final weights by $n_T$: $w_P = n_T ~ \tilde{w}_P$, $w_N = n_T ~  \tilde{w}_N$, and $w_R = n_T ~ \tilde{w}_R$. 
Detailed ablation studies on the three hyperparameters $C$, $V$, and $S$ are provided in the appendix.
We linearly scale the output of our model to $[-100,100]$ 
and display this relevance value for any paper on Scholar Inbox (\figref{fig:digest}).

\subsubsection{Solving the Cold Start Problem}
\begin{figure}[t!]
    \centering
    \includegraphics[width=0.6\textwidth]{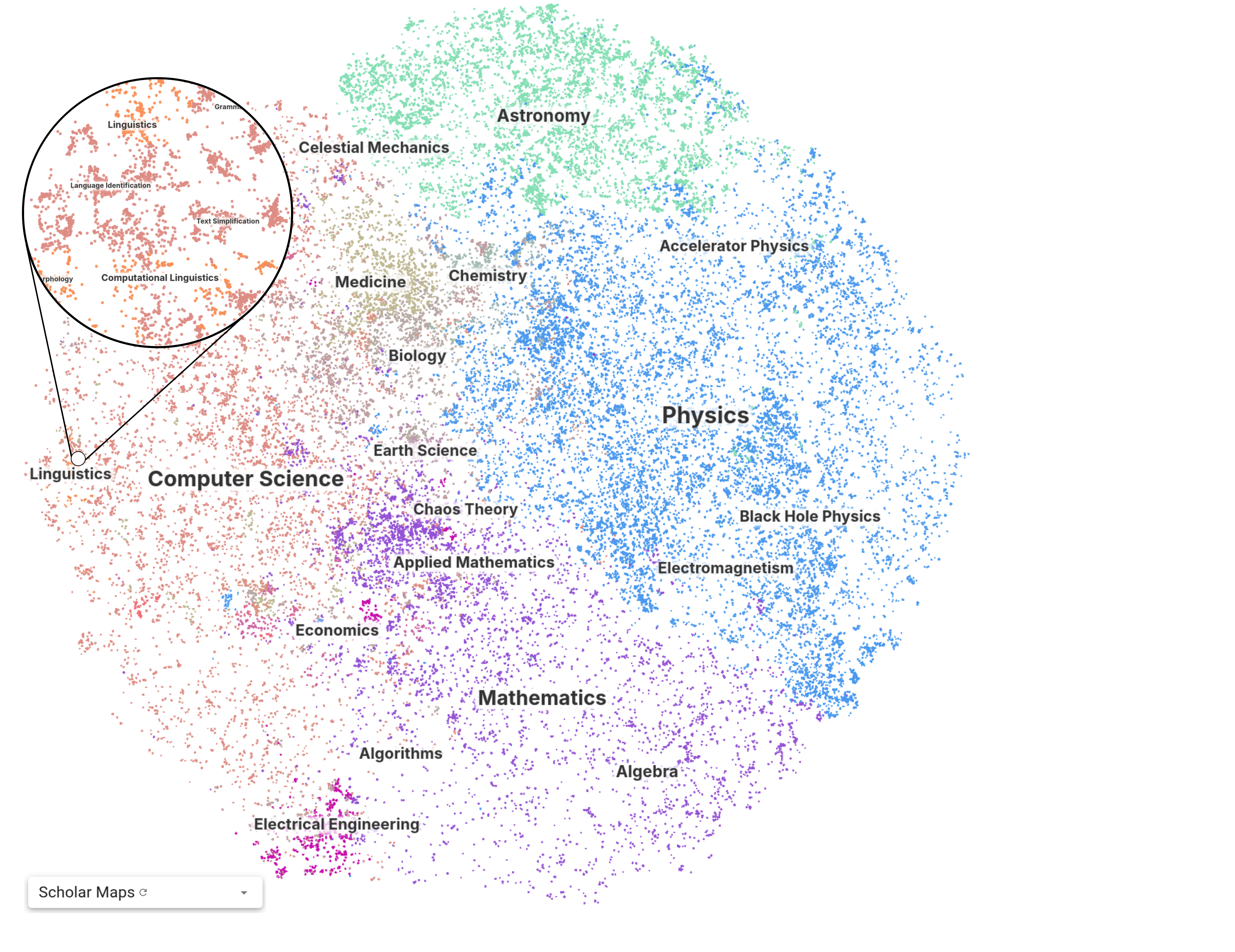}
    \caption{A t-SNE projection of the embedding space of all 3M papers in our database. 
    The most cited papers and biggest topics are shown first. As the user zooms in, 
    more papers are loaded dynamically.} 
    \label{fig:scholar_maps}
    \vspace{-2pt}
\end{figure}

The cold start problem of recommender systems consists of the lack of user interaction history for new users.
To provide an easy way to register to Scholar Inbox we
 offer users to add their own publications or publications from related authors via a simple author search.
 Alternatively, we allow users to navigate Scholar Maps, a 2D map of science, to quickly find relevant research fields and papers.
We show a screenshot of \url{scholar-maps.com} in \figref{fig:scholar_maps}.
The map is overlaid with topic labels, which we generated using Qwen \cite{qwen_qwen25_2025}. 
We provide the prompt engineering strategies for label generation in the appendix. 
Labels are generated for four hierarchy levels (field, subfield, subsubfield, method), 
such that the field (Computer Science, Physics, etc.) is shown on the outermost zoom level.
Subfields and method names of impactful papers are shown when zooming in, following Shneiderman's mantra "Overview first, zoom and filter, then details on demand" \cite{shneiderman_eyes_nodate}. 
Once users find their research area, they select multiple papers that they are interested in.\\
Users may search for papers by title or authors and add papers that they like to their selection. 
In a second step, we provide an active learning framework, 
which employs stratified sampling, prioritizing papers near and above the recommender's decision boundary, 
and prompts the user to rate them.
The recommender trains again after each rating is submitted, leading to iterative improvements.

\begin{table}[!t]
  \centering
  \resizebox{0.49\textwidth}{!}{%
  \small
  \begin{tabular}{l|cccccc|c}
  \toprule
    Key Features
       & \makecell{Google\\Scholar} 
       & \makecell{Semantic\\Scholar} 
       & \makecell{X} 
       & \makecell{Emati} 
       & \makecell{Arxiv\\Sanity} 
       & \makecell{Research\\Rabbit} 
       & \makecell{\textbf{Scholar}\\ \textbf{Inbox}} \\
      \midrule
      Daily Recom. & \xmark & \xmark & \checkmark & \xmark & \checkmark & \checkmark & \checkmark\\
      Multi-domain & \checkmark & \checkmark  & \checkmark  & \xmark & \xmark & \checkmark & \checkmark\\ 
      Non-redundant & \xmark & \checkmark & \xmark & \checkmark & \checkmark & \checkmark & \checkmark\\
      User ratings & \xmark & \xmark & \checkmark & \checkmark & \checkmark & \xmark & \checkmark\\
      Lexical search & \checkmark & \checkmark & \checkmark & \checkmark & \checkmark & \checkmark & \checkmark\\ 
      Semantic search & \checkmark & \xmark & \xmark & \xmark & \xmark & \xmark & \checkmark\\ 
      Collections  & \checkmark & \checkmark & \xmark & \xmark & \checkmark & \checkmark & \checkmark\\ 
      Paper maps & \xmark & \xmark & \xmark & \xmark & \xmark & \checkmark & \checkmark\\ 
      Dataset release  & \xmark & \xmark & \xmark & \xmark & \xmark & \xmark  & \checkmark\\ 
      \bottomrule
  \end{tabular}
  }
 \caption{Comparison of features across research recommendation platforms,  where \textit{Daily Recom.} denotes daily recommendation,
  \textit{User ratings} means the integration of user satisfaction metrics, 
  and \textit{Paper Maps} denotes the visualization of papers.}
  \label{tab:system-comparison}
\end{table}

\subsection{User Centric Design}
Most design decisions and features are first conceived by our users, before they are implemented by us. 
To reiterate the user focus, solicit user feedback and to make certain that Scholar Inbox addresses the concerns of its 
users, we regularly conduct user surveys.\\
As shown in \figref{fig:digest}, our website design follows a flat information hierarchy to minimise the number of clicks 
required to navigate to the desired functionality. 
The regular nature of our digest updates provides a habit forming experience, allowing our users to integrate 
Scholar Inbox into their daily work routine. 
We show a comparison of our features with other websites that recommend papers to researchers in \tabref{tab:system-comparison}.
\begin{figure}[t!]
    \centering
    \includegraphics[width=0.485\textwidth]{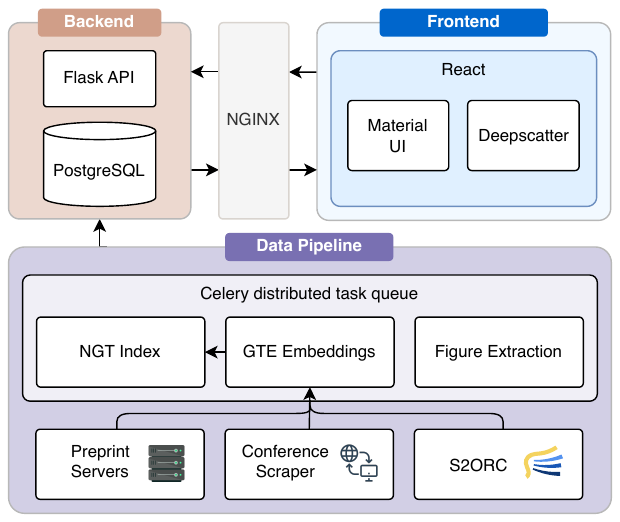}
    \caption{Data flow through our processing pipeline.} 
    \label{fig:software_architecture}
    \vspace{-2pt}
\end{figure}
\subsection{Software Architecture}
\figref{fig:software_architecture} shows the data processing pipeline. 
Scholar Inbox downloads papers and their metadata from preprint servers such as arXiv, bioRxiv, chemRxiv and medRxiv as well as directly 
from public conference proceedings. We compare and update missing fields in our database using 
the Semantic Scholar Open Research corpus (S2ORC) \cite{lo-wang-2020-s2orc}, 
to ensure that all papers are assigned the correct conference or journal upon publication. 
We also incorporate author information and the citation graph from S2ORC. 
We concatenate titles and abstracts, separated by a special [SEP] token, to encode each paper with $\text{GTE}_\text{large}$ \cite{li2023towards}, 
an efficient state-of-the-art transformer encoder trained with multi-stage contrastive learning. The paper embeddings are stored in  
NGT\footnote{\url{www.github.com/yahoojapan/NGT}}, a high performance nearest neighbor search index. 
We use Celery\footnote{\url{https://docs.celeryq.dev}} to handle asynchronous tasks, including extracting figures and text embeddings.
NGINX is used to serve the frontend static files and to proxy requests to the backend and
our user interface is built with React\footnote{\url{www.reactjs.org}}.
Scholar Maps uses deepscatter\footnote{\url{www.github.com/nomic-ai/deepscatter}} with tiled loading and GPU acceleration 
using WebGL to provide a smooth user experience.

\subsection{Daily Digest}
\begin{figure}[t!]
\centering
\includegraphics[width=0.49\textwidth]{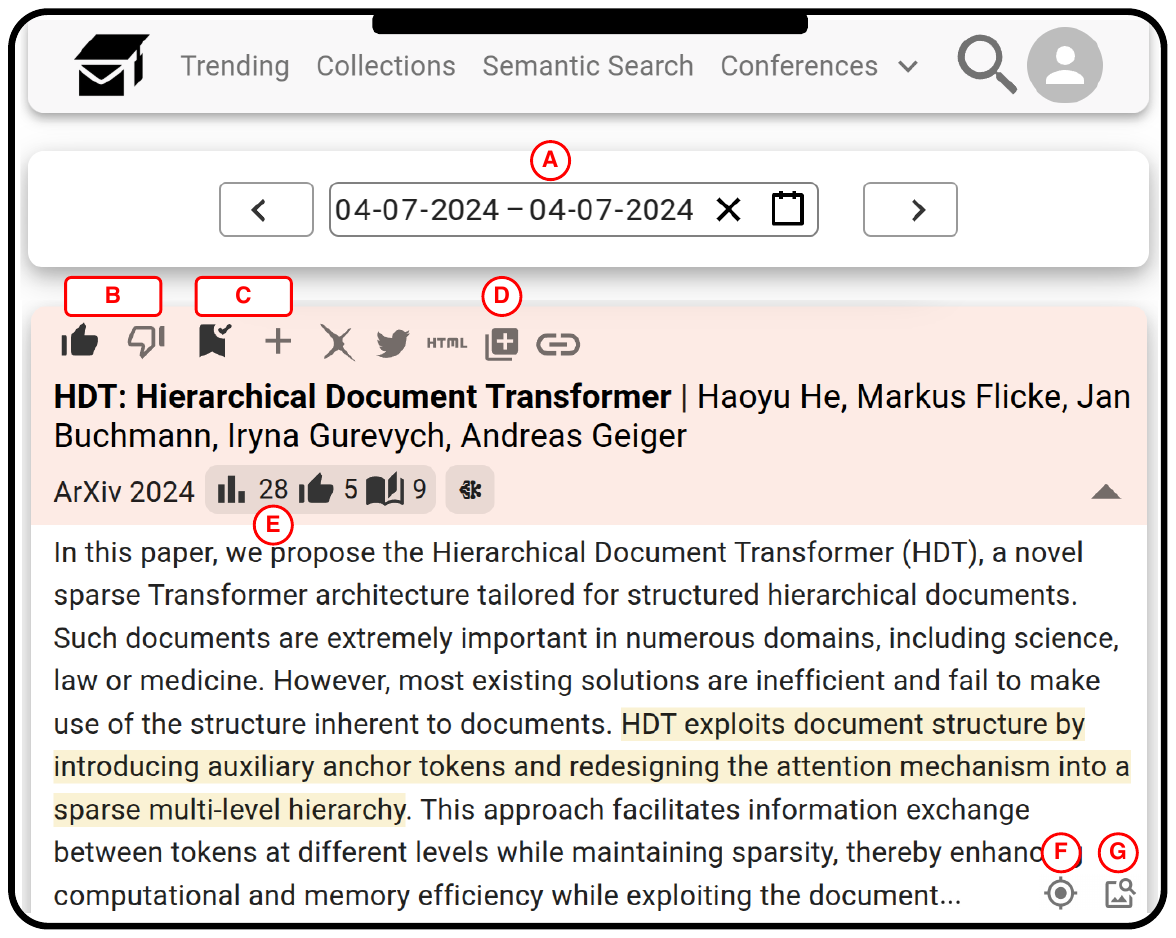}
\caption{Tablet or mobile phone view of the daily digest. 
To enable faster skim-reading, we highlight the sentence that is most related to the idea of the research paper. 
In red circles, we show the (A) date picker, (B) thumbs up/down buttons, 
(C) bookmarking/collections buttons, (D) bibtex button, (E) paper relevance score, 
(F) similar papers button and (G) teaser figure button.}
\label{fig:digest}
\vspace{-2pt}
\end{figure}

The daily digest, as shown in \figref{fig:digest}, is the main feature of Scholar Inbox. 
It holds a ranked list of papers within a short time period (day or week) with title, abstract, authors and publication venue for each paper. 
Digest papers are ordered by their predicted relevance for the current user, 
which also determines the paper header's background color.  
Users may refine their recommendation model by rating papers positively or negatively using the thumbs buttons (B).  
Using a button, each paper shows images of figures and tables, 
as well as the option to show a list of semantically similar papers. 
Moreover, users can search for semantically similar papers (F) and preview a paper's figures and tables (G) with a single click.
Papers can also be bookmarked or added to collections (C), posted on social media or exported as bibtex to 
reference managers (D). 
In addition to viewing daily digests, the user may also aggregate relevant papers over a longer time range (A) 
and specify the weekdays on which to receive their digests via email. 
If a user returns to the site after an extended period of time, 
we provide a catch-up digest containing the most relevant papers during their time of absence. 

\section{Evaluation}
\label{sec:evaluation}
\subsection{Recommendation model}
\begin{table}[h]
    \centering
    \renewcommand{\arraystretch}{1.3} 
    \resizebox{\columnwidth}{!}{%
    \fontsize{18}{18}\selectfont 
        \begin{tabular}{ll|cc|cc}
        \toprule
            Model & Dim. & F1 & nDCG & Balanced acc.  & AUC \\
            \midrule
            TF-IDF & 10k & 83.60 ${\scriptstyle \pm 0.10}$ & \textbf{88.67} ${\scriptstyle \pm 0.29}$ & 75.74 ${\scriptstyle \pm 0.05}$ & 84.41 ${\scriptstyle \pm 0.09}$ \\
            TF-IDF & 256 & 81.03 ${\scriptstyle \pm 0.17}$ & 83.37 ${\scriptstyle \pm 0.26}$ & 74.52 ${\scriptstyle \pm 0.10}$ & 82.28 ${\scriptstyle \pm 0.04}$ \\
            SPEC2 & 256 & 83.22 ${\scriptstyle \pm 0.16}$ & 84.21 ${\scriptstyle \pm 0.31}$ & 78.16 ${\scriptstyle \pm 0.07}$ & 86.36 ${ \scriptstyle \pm 0.09}$ \\
            GTE-B & 256 & 84.16 ${\scriptstyle \pm 0.11}$ & 85.42 ${\scriptstyle \pm 0.28}$ & 77.92 ${\scriptstyle \pm 0.08}$ & 86.24 ${\scriptstyle \pm 0.05}$ \\
            GTE-L & 256 & \textbf{84.51} ${\scriptstyle \pm 0.15}$ & 85.83 ${\scriptstyle \pm 0.22}$ & \textbf{78.31} ${\scriptstyle \pm 0.12}$ & \textbf{86.75} ${ \scriptstyle \pm 0.07}$ \\
            \bottomrule
        \end{tabular}
}
    \caption{Performance of the recommender using different embedding methods. 
    TF-IDF 10k is sparse with 10K dimensions, 
    while the other models are dense and compressed to 256 dimensions using PCA.}
    \label{tab:recommender_results}
\end{table}
\noindent
We evaluate classic sparse (TF-IDF) and neural network-based dense (GTE, SPECTER2) embedding models for encoding research papers, 
measuring performance through two distinct approaches in \tabref{tab:recommender_results}. 
First, we follow established methodologies for recommender systems without explicit negative ratings \cite{he-etal-2017-ncf}
and evaluate each positive sample together with randomly sampled negative examples. 
For these, we compute F1-score and nDCG using a leave-one-out strategy for positively voted validation papers.
While this evaluation is widely adopted in the literature, it fails to account for the impact of hard negative samples.
We further analyze model performance including explicit negative user ratings on binary classification metrics (balanced accuracy and AUC)
and find that GTE outperforms TF-IDF on classification between positives and hard negatives.
\begin{figure}[t!]
    \centering
    \includegraphics[width=0.485\textwidth]{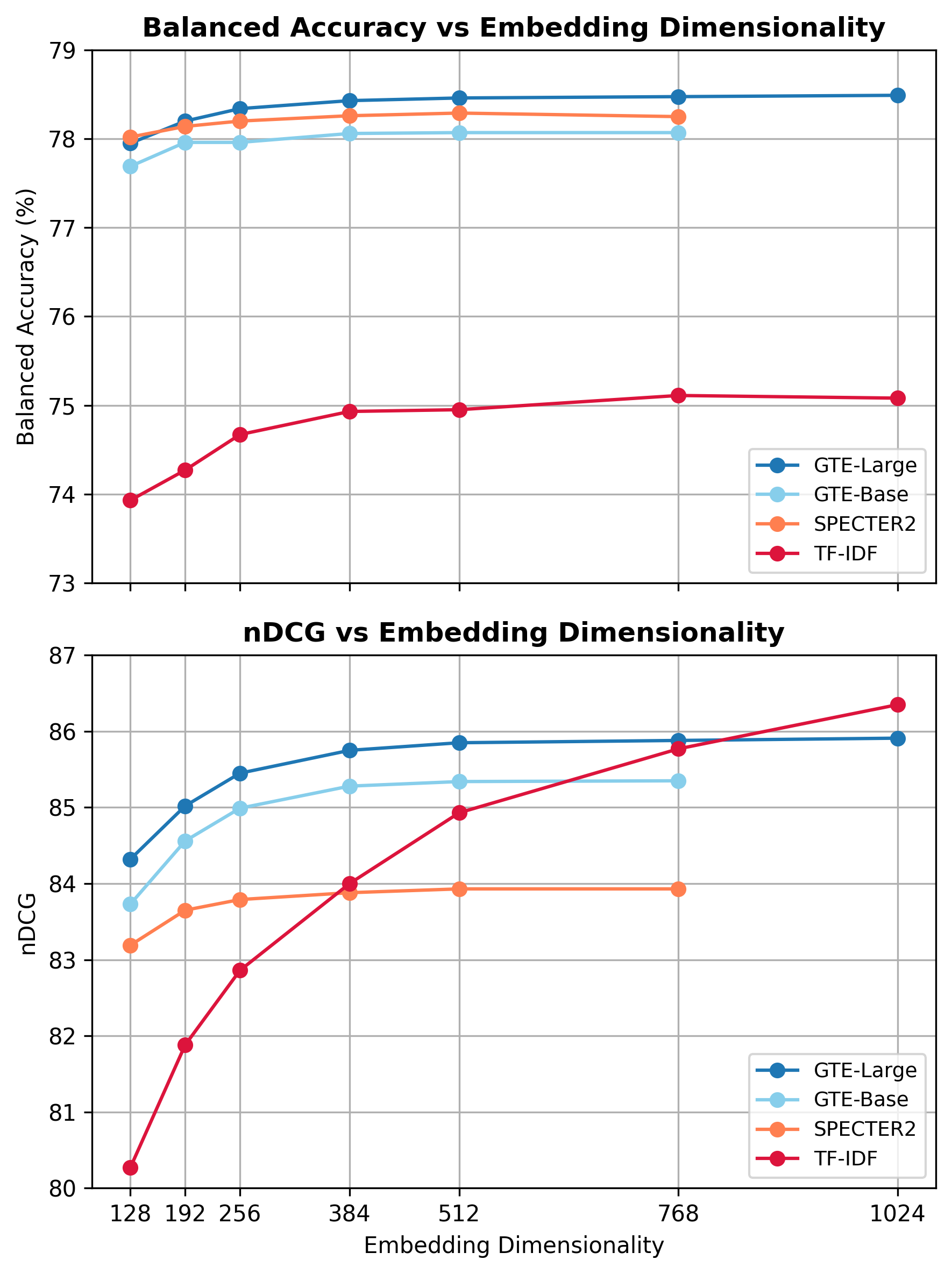}
    \caption{
    Performance of different embeddings after dimensionality reduction from their original sizes: GTE(1024), SPECTER(768), 
    TF-IDF(10k). At its orginial dimensionality of 10k, TF-IDF achieves a score of 88.2 on nDCG.}
    \label{fig:embeddings_pca}
\end{figure}

\begin{figure}[t!]
    \centering
    \includegraphics[width=0.49\textwidth]{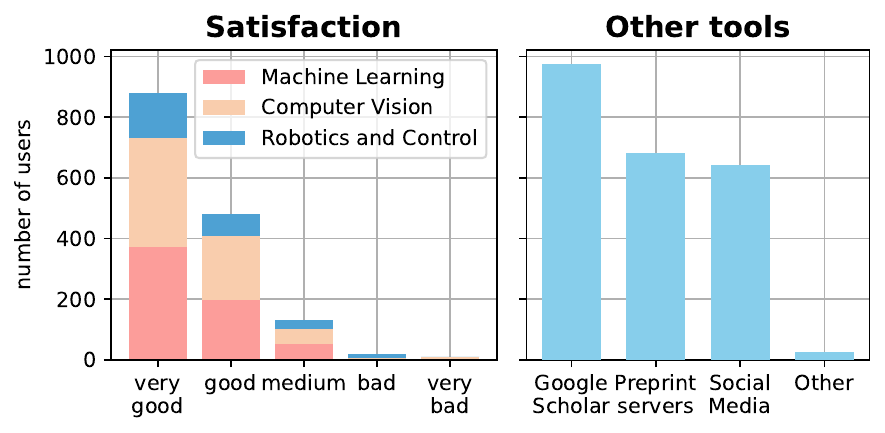}
    \caption{User study for Scholar Inbox among 1,233 participants. Left: Satisfaction levels across users in Machine Learning, Computer Vision, and Robotics indicate a highly positive experience. Right: Users also use search engines, preprint servers, and social media, but few rely on other recommender systems, underscoring Scholar Inbox’s central role in paper discovery.}
    \label{fig:satisfaction}
\end{figure}

\begin{figure*}[t!]
    \centering
    \includegraphics[width=0.99\textwidth]{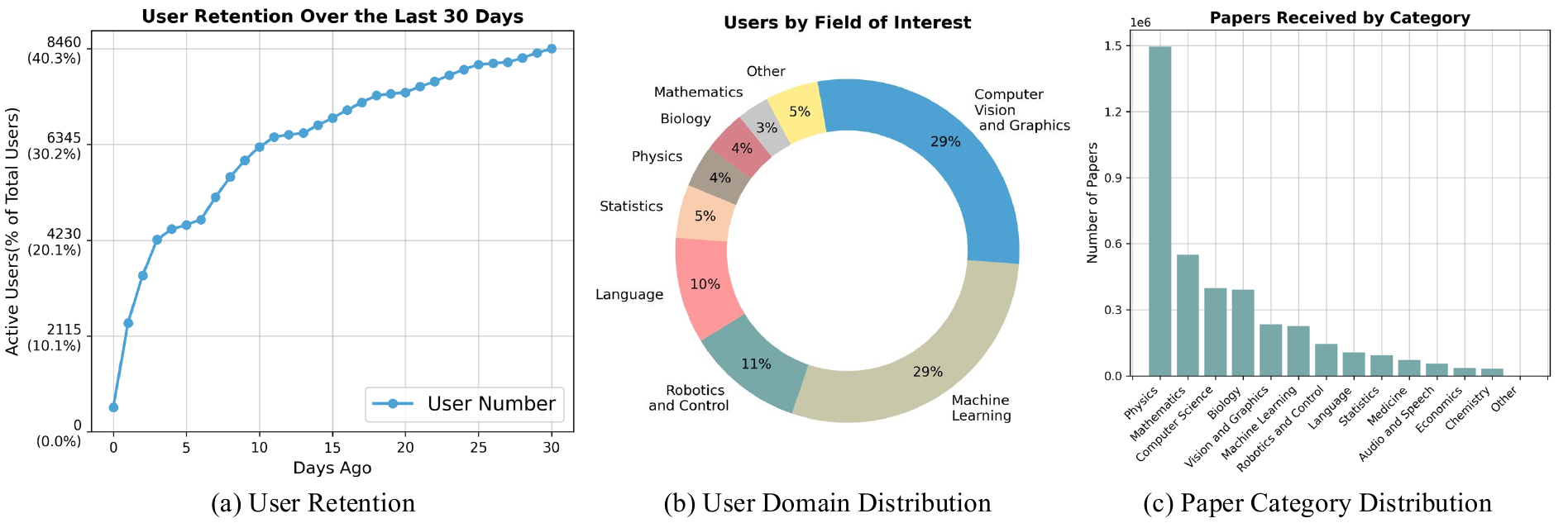}
    \caption{Statistics of user and papers in Scholar Inbox: (a) Cumulative number of active users in the past 30 days, demonstrating a high retention rate; (b) Distribution of users by research domain, indicating strong representation in ML and CV while maintaining multidisciplinary reach; and (c) Distribution of recommended papers by category, reflecting user interests across diverse scientific fields.}
    \label{fig:more_statistics_analysis}
\end{figure*}

Evaluating qualitatively, we find GTE underperforms on nDCG primarily for two reasons: 
It assigns higher probabilities to sampled negatives that 
resemble users' positive training examples, 
and it assigns lower probabilities to certain positive validation papers 
which are also classified negatively by TF-IDF.
 The first case is susceptible to noise and 
 the second has minimal impact on the digest, as neither model recommends these false negatives. 
 Therefore, we select the GTE-Large model for its superior performance on explicit user ratings, 
 which we consider more reliable. 
 Empirically, we also find that our dense embeddings yield better calibrated cosine similarities which benefit similar 
 paper/semantic search and 2D visualizations like Scholar Maps.
\\\\

Step further, we investigate the effect of PCA-based dimensionality reduction on transformer-derived embeddings for the recommendation task, as shown in \figref{fig:embeddings_pca}.
Performance is evaluated in terms of balanced accuracy (top) and nDCG (bottom) across varying embedding dimensions. For all transformer-based methods (GTE-Large, GTE-Base, and SPECTER2), both metrics remain relatively stable when reducing dimensions from 1024 to 256, suggesting redundancy in the original representations. Below 256, however, a notable degradation in performance emerges, indicating that further compression removes informative components. We conclude that not all dimensions are used efficiently for our recommendation task. Notably, TF-IDF exhibits steady gains with increased dimensionality.
For runtime and memory efficiency 
we choose a dimensionality of 256 for the final GTE-large model.

\subsection{User Study}
To evaluate Scholar Inbox, we conduct a user study with 1233 participants, 
who are asked to rate their satisfaction with the platform on a scale from 1 to 5 in terms of usability, satisfaction, 
and the quality of recommendations. 
Their evaluation of Scholar Inbox is extremely positive, as can be seen from user voting in \figref{fig:satisfaction} 
and from the user retention statistics in \figref{fig:more_statistics_analysis}(a). 
The most common criticism from our user study is that the platform currently does not support explicit modeling of separate research interests. 
Whilst we observe that multiple research interests are already handled well in a single recommender, we are working on enabling users to explicitly switch 
between different research interests in the next version of Scholar Inbox.

\subsection{User Retention and Distribution}
In \figref{fig:more_statistics_analysis}(a), we present the cumulative number of users active in the last 30 days. 
This graph only shows user interactions on the website, excluding users who only read our email newsletter.
Even though the number of registered users on Scholar Inbox is only 23k, which is relatively few for a website, 8k (35\%) of them were active in the last 30 days. 
This high retention rate underscores both the effectiveness of the recommendation system and the practical value offered by the platform.

As shown in \figref{fig:more_statistics_analysis}(b), while a significant portion of users focus on Machine Learning and Computer Vision, the presence of users from diverse fields such as Physics, Biology, Language, and Statistics demonstrates that our platform is attracting a broad range of researchers. This suggests its potential to effectively support interdisciplinary research across multiple scientific domains.



\section{Conclusion}
Scholar Inbox is a new open-access platform that provides daily, personalized recommendations for research papers 
and a range of tools to improve research workflows and promote open access to research. 
Our evaluation on a dataset of 800k user ratings and the user study
 highlight the platform's effectiveness in providing accurate recommendations and enhancing user satisfaction.

\subsection*{Acknowledgements}
Andreas Geiger is a member of the Machine Learning Cluster of Excellence, funded by the
Deutsche Forschungsgemeinschaft (DFG, German Research Foundation) under Germany’s
Excellence Strategy – EXC number 2064/1 – Project number 390727645. 
Bora Kargi and Kavyanjali Agnihotri were funded by the ELIZA master's scholarship. This project was supported by a VolkswagenStiftung Momentum grant.
\bibliography{custom}

\clearpage
\section{Appendix}

\subsection{Prompt Engineering Strategies for t-SNE Label Generation}
To extract the topic hierarchy for t-SNE visualization, we conducted LLM inference on each paper using a 
prompt composed of four distinct parts: Task, Additional Note, Format, and Title \& Abstract. 
The Task section provides the general extraction instructions and mandates strict adherence to 
the specified format while explicitly instructing the model to omit any additional commentary 
to simplify output parsing. The Additional Note section restricts the field values to a predefined, 
handcrafted list of scientific disciplines. The Format section details the precise structure of the 
expected output along with explanations of the corresponding fields. Finally, the Title \& Abstract 
section contains the actual text to be processed for extracting the required information.

During prompt engineering, we determined that including the format explanation only once, 
positioned as late as possible before the data, is optimal. Moreover, employing an explicit empty 
field placeholder proved crucial for smaller LLMs, as it enhances structural consistency 
and prevents unnecessary repetitions in the output.

\begin{lstlisting}[style=promptstyle,caption={Scholar Map's label generation prompt. For better readability, we shortened the list of disciplines.}]
Task: Based on the title and abstract provided, extract 
and label the following key details exactly as specified:
field_of_Paper, subfield, sub_subfield, keywords, method_
name_shortname. Follow the structure exactly and keep your 
answers brief and specific. Adhere strictly to the format. 
If any information is unclear or unavailable in the abstract, 
write "None." for that field. Use the exact labels and 
formatting provided. Do not include comments or repeat any 
part of the response. Note: For field_of_Paper, choose one 
from the following list of academic disciplines: 
Mathematics, Physics, Chemistry, ...

Details to Extract:
field_of_Paper = 
*The primary academic discipline from the list above.* 
[insert answer]
subfield = 
*The main research category within the field.* 
[insert answer]
sub_subfield = 
*A narrower focus within the subfield.* 
[insert answer]
keywords = 
*A set of 3-5 words or phrases that describe the core topics, 
separated by commas.* 
[insert answer]
method_name_shortname = 
*The main technique or model name proposed in the abstract.* 
[insert answer]

Title: [title]
Abstract: [abstract]
\end{lstlisting}

\subsection{Technical Challenges}
Extracting teaser figures (or getting GTE embeddings) is compute-intensive; 
however, leveraging GPU acceleration facilitates rapid inference and efficient parallel processing of papers. 
For efficiency our architecture enables external machines to connect to the main server's broker and backend 
(powered by Redis) via SSH port forwarding. This setup allows remote Celery workers to access tasks directly from the Scholar server. 
Consequently, any machine with the appropriate credentials—regardless of its physical location—can 
serve as a task consumer within our distributed environment, making our pipeline scalable by allowing us to 
seamlessly connect additional machines to accelerate computations as needed.


\begin{figure*}[ht]
    \centering
    \includegraphics[width=1.0\textwidth]{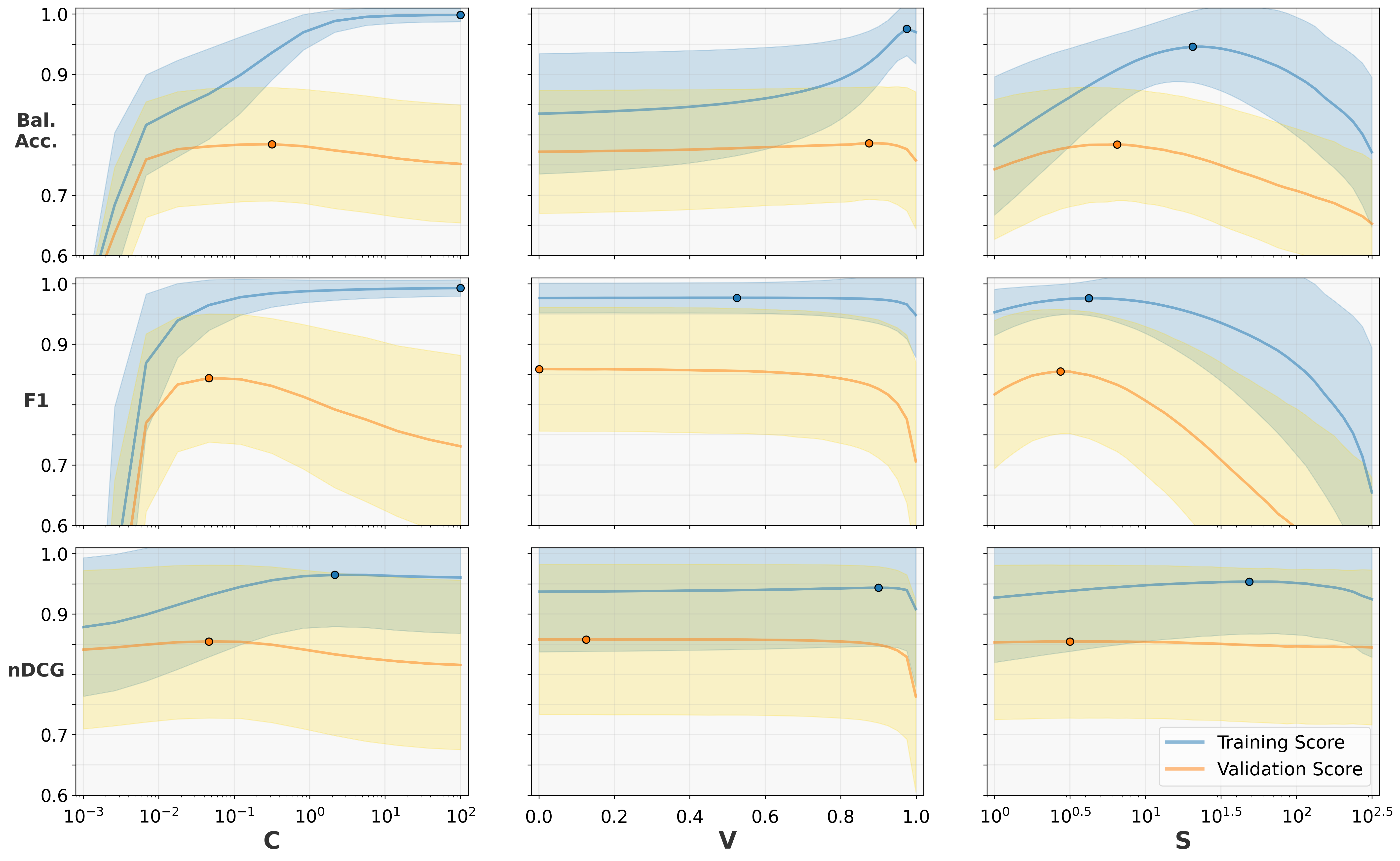}
    \caption{Hyperparameter ablation studies on GTE-Large embeddings. The metrics correspond to those in Table \ref{tab:recommender_results}. Each plot shows the effect of individually varying one parameter while keeping the others fixed. Shaded regions indicate ± 1 standard deviation across the user base (not across random seeds).}
    \label{fig:ablation_study}
\end{figure*}

\subsection{Hyperparameter Ablation Studies}
We evaluate the sensitivity of our system to each of the three hyperparameters introduced in Section~\ref{subsubsec:training}. For our ablation experiments, we use 256-dimensional GTE-Large embeddings with a standard configuration of $(C = 0.1, V = 0.8, S = 5.0)$. As in our main evaluation, balanced accuracy is calculated using explicit negative votes, while F1 and nDCG refer to 100 randomly sampled negatives. The results are summarized in Figure~\ref{fig:ablation_study}.

\subsubsection{Inverse Regularization Strength C}
With $V$ and $S$ fixed at their standard configuration values, positive weights $w_P$ are higher than negative weights $w_N$ and $w_R$. The model prioritizes fitting positive training examples, achieving the highest recall at $C=10^{-1.5}$ (where F1 and nDCG are maximized). Further increasing $C$ allows the model to better fit explicit negative examples, improving specificity and balanced accuracy (optimal at $C=10^{-0.5}$). However, this tightens the decision boundary around difficult negatives, reducing performance between positives and simpler sampled negatives, consequently lowering F1 and nDCG. We note that linear classification applied to higher-dimensional embeddings contains a larger number of parameters and therefore attains similar performance under stronger regularization (e.g. $C = 0.05$ for 1024-dimensional GTE-Large).

\subsubsection{Explicit-to-Random Negative Ratio V}
The hyperparameter $V$ controls the trade-off between performance on explicit negatives and randomly sampled negatives. Raising it from $0$ to $0.9$ elevates specificity on explicit negatives from $68\%$ to $78\%$ and maximizes balanced accuracy at $78.6\%$ (up from $77.2\%$). The increased emphasis on difficult negative examples again tightens the decision boundary, producing false negatives and causing a monotonic decrease in F1 and nDCG. Nonetheless, we select a larger value $V = 0.8$ as it makes the model more receptive to downvotes and allows users to tune their classifier by explicitly stating which papers should not be recommended to them. 

\subsubsection{Negative Weights Scale S}
The hyperparameter $S$ controls the magnitude of the negative weights $w_N$ and $w_R$. At low values ($S=1$), the model exhibits highly imbalanced class behavior with a recall of 94\% but a specificity on explicit negatives of only 55\%. Raising $S$ mitigates this disparity, with all three metrics reaching high scores at our standard configuration value.
As $S$ increases, the model assigns progressively lower logits to all samples. Beyond $S=5$, this reduction becomes substantial enough to cause a notable 
drop in recall, lowering balanced accuracy and F1. In contrast, nDCG remains stable up to much higher values ($S=10^3$) because the model preserves the relative ranking between positives and randomly sampled negatives until positive weights become negligibly small compared to negative weights.

\end{document}